\documentclass[twocolumn]{bytedance_seed}

\usepackage[toc,page,header]{appendix}
\usepackage[table]{xcolor}
\usepackage{multirow}
\usepackage{pifont}
\usepackage{makecell}
\usepackage{enumitem}

\usepackage{minitoc}
\usepackage{amssymb}
\usepackage{multirow}
\usepackage{makecell}

\title{DanceHMR: Hand-Aware Whole-Body Human Mesh Recovery from Monocular Videos}

\author[*]{Wenhao Shen}
\author[*,\dagger]{\ \ Ming Zhou}
\author[]{Hengyuan Zhang}
\author[]{Siyuan Bian}
\author[]{\\ Youjiang Xu}
\author[]{Yuan Zhang}

\affiliation{ByteDance Intelligent Creation}

\contribution[*]{Equal contribution}
\contribution[\dagger]{Corresponding author}

\abstract{
Monocular video human mesh recovery is essential for digital humans, avatar animation, and embodied simulation, where both temporal stability and expressive whole-body motion are required. Existing video HMR methods produce coherent body motion but often overlook detailed hand articulation, while image-based whole-body methods recover SMPL-X meshes independently per frame, often leading to jittery and inaccurate hand motion. We present a temporally coherent whole-body HMR framework for challenging in-the-wild monocular videos. Our model unifies body context and part-specific hand observations through residual body-hand fusion, enabling stable body motion and detailed hand recovery within a single temporal architecture. We further introduce close-up-aware augmentation to improve robustness under upper-body framing. Experiments on whole-body and body-only benchmarks demonstrate improved hand reconstruction and competitive body accuracy. Our method also produces temporally stable and 2D-consistent SMPL-X motion in challenging real-world videos.
}

\date{\today}
\correspondence{Ming Zhou at \email{zhouming.9527@bytedance.com}}

\checkdata[Project Page]{\url{https://shenwenhao01.github.io/dancehmr/}}

\begin{document}
\maketitle

\begin{figure*}[htbp]
    \centering
    \vspace{-15pt}
    \includegraphics[width=1.0\textwidth]{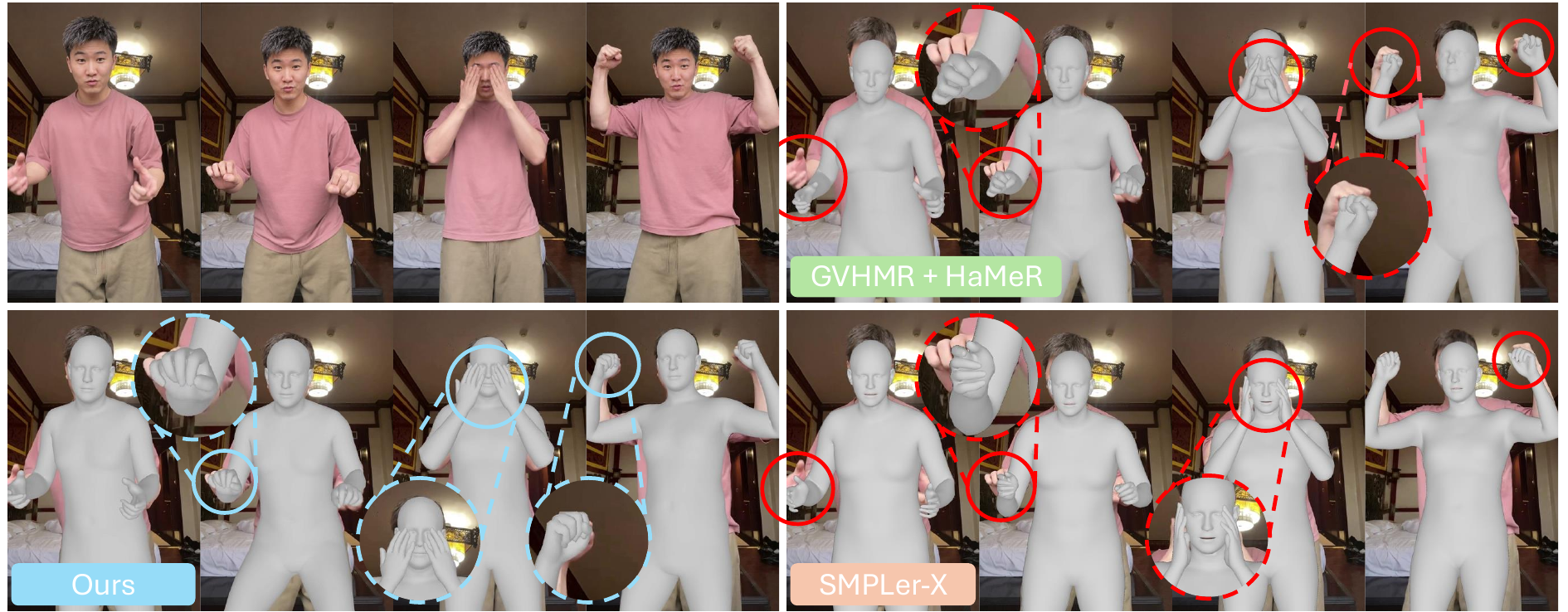}
    \caption{Given a monocular video, our method reconstructs temporally coherent and 2D consistent whole-body motion. Previous video-based HMR methods focus predominantly on the body; when combined with hand models (e.g., GVHMR~\cite{shen2024gvhmr} + HaMeR~\cite{pavlakos2024hamer}), they often produce unnatural wrist orientations and temporal inconsistencies. Image-based whole-body models such as SMPLer-X~\cite{cai2024smplerx} process frames independently and struggle with small hand crops, resulting in inaccurate and jittery hand motion. In contrast, by modeling the body and hands together, our approach produces temporally stable hand articulation and fluid whole-body dynamics.}
    \label{fig:teaser}
    \vspace{-10pt}
\end{figure*}

\newpage

\section{Introduction}
\label{sec:intro}
Monocular human mesh recovery (HMR) from videos is a foundational technology for scalable human digitization, serving industrial applications such as motion generation data acquisition, avatar animation, and embodied AI. For these downstream systems, temporal coherence is essential; frame-wise jitter or physical discontinuity renders the extracted motion unusable for training pipelines or physical simulation. Furthermore, in real-world scenarios like livestreaming, dancing, and human-object interaction, hands are central to conveying motion semantics and visual realism. Recovering accurate, expressive, and temporally stable articulations for both the body and hands jointly is therefore a critical challenge.

Recent video HMR methods~\cite{shin2024wham,shen2024gvhmr, yin2024whac, chen2025human3r, genmo2025,wang2026duomo} stabilize body motion but typically neglect detailed hand articulation. 
Therefore, they are insufficient for close-up and hand-centric videos where precise finger motion and wrist-body consistency are important.
A straightforward solution is to combine a video-based body estimator with a hand reconstruction model. 
However, the independent estimation of the body and hands usually leads to physically disjointed wrist orientations.
Moreover, hand self-occlusion, blur and partial visibility are common in real-world videos, making image-based hand predictions prone to severe jitter and implausible poses.

Another challenge comes from the camera composition of real-world videos. Many internet videos are not captured as standard full-body sequences found in typical motion capture benchmarks. Livestream, vlog, and speech videos heavily feature close-up or upper-body framing, with hands frequently moving in and out of the image boundaries. Existing models trained predominantly on full-body crops suffer from a severe distribution gap in these scenarios, leading to inaccurate body-to-image alignment, distorted head projections, and unstable pose estimates.

These observations motivate a simple but important design principle: body motion, hand articulation, and temporal context should be modeled jointly rather than independently. The body pose provides strong spatial and anatomical constraints for wrist orientation and general hand placement, while focused hand crops and 2D keypoints provide complementary local evidence for complex finger articulation. Furthermore, temporal context is essential to infer plausible hand poses when visual observations of the extremities are noisy, blurred, truncated, or temporarily missing.

In this paper, we present a temporally coherent whole-body HMR framework designed for challenging in-the-wild monocular videos. Given a video sequence, our model jointly predicts body and hand parameters in a unified temporal architecture. It integrates part-specific observations through a hand-aware temporal transformer. 
Instead of directly stitching hard pose constraints from an image-based hand model, we encode hand visual features and explicit 2D hand geometry with independent left- and right-hand branches. The encoded hand evidence is then injected as residual refinements into a body-anchored temporal representation.
This acts as a set of complementary soft observations, allowing the model to adaptively utilize reliable visual cues when hands are visible, and rely on temporal context and geometric priors when they are occluded.

To bridge the gap between benchmark-style full-body data and real-world close-up videos, we introduce a truncation-aware training strategy. By simulating temporally consistent half-body crops from existing datasets, we expose the network to diverse partial-body framing and hand entry/exit events. We train the model in a two-stage curriculum: first learning broad temporal body and hand motion priors from mixed-quality data, and subsequently emphasizing high-quality hand data and distal fingertip supervision to refine fine-grained hand articulation. 

In summary, our main contributions are:
\begin{itemize}
    \item We propose DanceHMR, a temporally coherent whole-body HMR framework that recovers stable body and hand motion from monocular videos.
    \item We propose a close-up-aware augmentation strategy for upper-body and truncated video compositions, improving robustness in challenging real-world scenarios such as vlog, livestream, and hand-centric videos.
    \item Extensive experiments demonstrate that our method substantially improves hand reconstruction accuracy and temporal stability over existing whole-body methods.
\end{itemize}

\section{Related work}

\subsection{Image-based whole-body and hand mesh recovery}

Recent image-based human mesh recovery methods extend monocular HMR to expressive SMPL-X estimation.
ExPose~\cite{choutas2020expose} and PIXIE~\cite{feng2021pixie} combine part-specific body, hand, and face experts to improve local reconstruction quality.
PyMAF-X~\cite{zhang2023pymafx} further refines full-body alignment through mesh-aligned feedback.
More recent methods~\cite{lin2023osx, cai2024smplerx, hmradapter, wanqiyin2026smplestx} further improve large-scale expressive human reconstruction with stronger transformer backbones, scalable training strategies, or adaptation mechanisms.
These methods achieve impressive per-frame reconstruction quality and serve as strong baselines for body and hand mesh recovery.
However, they are primarily designed for individual images and therefore do not explicitly enforce temporal consistency when applied frame-by-frame to videos.

Beyond whole-body mesh recovery, recent works also focus on monocular hand reconstruction. 
HaMeR~\cite{pavlakos2024hamer} reconstructs detailed 3D hand meshes with transformer-based image modeling, while WiLoR~\cite{potamias2025wilor} further improves in-the-wild hand localization and reconstruction robustness.
More recent methods such as HaWoR~\cite{zhang2025hawor} extend hand reconstruction toward world-referenced hand motion recovery from monocular videos.
These methods achieve strong per-frame hand reconstruction quality when hands are clearly visible.
However, directly stitching independently estimated hands onto a video body sequence often leads to unstable hand motion and wrist inconsistency.
In contrast, our method jointly models body and hand motion in a unified temporal framework.
\subsection{Video human mesh recovery}

Video HMR methods address the temporal inconsistency of image-based regressors by exploiting motion context across frames. 
Early video HMR methods~\cite{kocabas2020vibe, choi2021tcmr, wei2022mpsnet, rempe2021humor, shen2023glot} mainly improve temporal body consistency through temporal dynamics modeling and motion-prior regularization. Later works~\cite{yuan2022glamr, ye2023slahmr, goel2023hmr2.0, sun2023trace} extend this line toward video-level tracking and global motion recovery under dynamic cameras.
However, their output space and evaluation focus are still largely body-centric. They mainly target SMPL body motion rather than detailed hand articulation, wrist-body consistency, or robust body and hand recovery in close-up and truncated videos.

\begin{figure*}[h]
  \centering
  \includegraphics[width=1.0\linewidth]{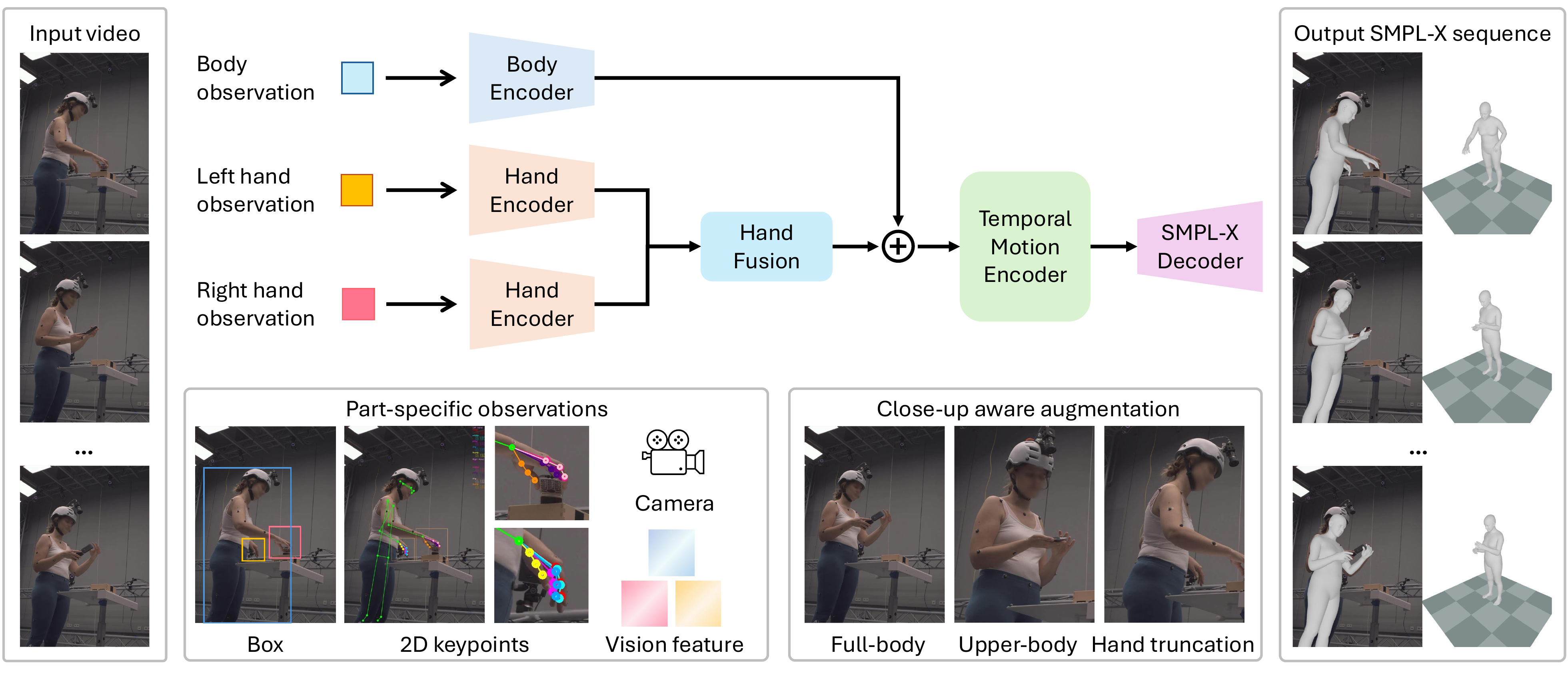}
  \vspace{-15pt}
  \caption{Framework overview. Given an input video, we separately encode part-specific body and hand observations.  Hand evidence is fused into the body representation as residual cues before temporal modeling. A temporal motion encoder then reconstructs temporally coherent SMPL-X motion sequences. During training, we additionally employ close-up-aware augmentation to improve robustness to upper-body views and hand truncation scenarios.}
  \label{fig:method}
  \vspace{-10pt}
\end{figure*}

Another line of work improves world-grounded video HMR.
WHAM~\cite{shin2024wham} introduces a body motion recovery framework that combines video features, camera motion, and contact-aware refinement.
TRAM~\cite{wang2024tram} estimates global human trajectory and local body motion by combining SLAM-based camera motion estimation with a video transformer.
GVHMR~\cite{shen2024gvhmr} further formulates global motion recovery in gravity-view coordinates to reduce camera-motion ambiguity.
WHAC~\cite{yin2024whac} and Human3R~\cite{chen2025human3r} also explore expressive or scene-aware video reconstruction in world coordinates.
In parallel, PromptHMR-Vid~\cite{wang2025prompthmr}, DUOMO~\cite{wang2026duomo}, and GENMO~\cite{genmo2025} investigate promptable or generative formulations for video human motion recovery, introducing stronger temporal and motion priors for ambiguous observations.
These methods significantly advance video-based human motion recovery, especially in temporal stability and global trajectory estimation.
Despite these advances, existing video HMR methods still mainly focus on body motion and global reconstruction, while detailed and temporally coherent hand recovery remains under-explored.
In contrast, our work focuses on temporally coherent body and hand recovery in challenging real-world videos with frequent hand occlusion and truncation.

\section{Method}

\subsection{Method overview}

Given a monocular video sequence $\mathcal{I}_{1:T}=\{I_t\}_{t=1}^{T}$, our goal is to recover a temporally coherent SMPL-X motion sequence including body pose, hand articulation, camera-space translation, gravity-aware global motion, and contact states. For each frame, we extract frame-aligned observations $\mathcal{C}_t = \{\mathcal{C}_t^b, \mathcal{C}_t^{lh}, \mathcal{C}_t^{rh}\}$. 
The body observation 
$\mathcal{C}_t^b = \{K_t^b, F_t^b, B_t^b, \Omega_t\}$ 
consists of body 2D keypoints $K_t^b$, body crop features $F_t^b$, a camera-aware body-box token $B_t^b$, and camera angular velocity $\Omega_t$. 
Similarly, the left- and right-hand observations 
$\mathcal{C}_t^{lh}=\{K_t^{lh},F_t^{lh},B_t^{lh}\}$ 
and 
$\mathcal{C}_t^{rh}=\{K_t^{rh},F_t^{rh},B_t^{rh}\}$ 
contain hand 2D keypoints, hand crop features, and camera-aware hand-box tokens.

Our architecture follows a body-centric but hand-aware design. We first encode body observations into a stable per-frame token. 
Left and right hand observations are encoded separately as part-specific hand evidence and fused into a residual hand vector, which is injected into the body token before temporal modeling. 
The fused token sequence is processed by a temporal motion encoder and decoded into a coupled camera-gravity SMPL-X representation $\mathbf{x}_{1:T}$. 
To improve robustness in real-world videos, we further introduce a close-up-aware training strategy that exposes the model to partially visible body and hand observations, such as close-up views and hands entering or leaving the image.
During inference, we optionally apply a physically grounded trajectory refinement step to reduce foot sliding and stabilize global motion.

\subsubsection{Camera-gravity SMPL-X representation}

Existing video HMR methods commonly represent human motion in the camera coordinate system, which facilitates image alignment and 2D reprojection supervision. 
However, relying solely on camera-space motion makes it difficult to maintain temporally stable motion under moving cameras. 
Following the gravity-view coordinate system~\cite{shen2024gvhmr}, 
we adopt a coupled representation that jointly models camera-aligned human motion and gravity-aware motion cues. 
Different from previous body-centric representations, we extend the representation to SMPL-X and explicitly model articulated hands.

Specifically, for each frame $t$, the global motion component consists of the gravity-view orientation 
$\Gamma_t^{GV} \in \mathbb{R}^{6}$ and the local root velocity 
$\mathbf{v}_t^{root} \in \mathbb{R}^{3}$. 
The local human pose is represented by the full SMPL-X pose rotations 
$\theta_t^{x} \in \mathbb{R}^{55 \times 3}$, which include body joints, hands, jaw, and eye poses. 
We also include the shape parameters $\beta_t \in \mathbb{R}^{10}$. 
To preserve image-space alignment, we include the camera-space human trajectory, consisting of the camera-frame orientation 
$\Gamma_t^{c}$ and translation 
$\tau_t^{c}$. 
Additionally, we predict static-contact states 
$\mathbf{p}_t$ for selected foot and wrist joints, which provide cues for global trajectory refinement.

The complete motion sequence is denoted as 
$\mathbf{x}_{1:T}=\{\mathbf{x}_t\}_{t=1}^{T}$, where each frame-level motion representation is defined as
\begin{equation}
\mathbf{x}_t =
\left(
\Gamma_t^{GV},
\mathbf{v}_t^{root},
\theta_t^{x},
\beta_t,
\Gamma_t^{c},
\tau_t^{c},
\mathbf{p}_t
\right).
\end{equation}
This representation combines camera-aligned reconstruction with gravity-aware motion cues. 
The camera-space components enable direct supervision from image observations, while the gravity-view orientation and root velocity improve temporal motion stability under moving cameras.
By explicitly modeling articulated hands in SMPL-X, our model preserves wrist-hand consistency.

\subsection{Hand-aware whole-body motion model}

\subsubsection{Body-anchored frame tokens}

We first construct a body-centric frame token to aggregate the global human context. To achieve this, we combine explicit geometric constraints from 2D keypoints with the appearance features of the body crop. We also incorporate a camera-aware bounding box token to anchor the subject's scale and location within the full image, alongside camera angular velocity to decouple human movement from camera ego-motion. These distinct observation streams are projected into a shared latent space and fused additively to form the body token $z_t^b$:
\begin{equation}
z_t^b = E_{K}^{b}(K_t^b) + E_{F}^{b}(F_t^b) + E_{B}^{b}(B_t^b) + E_{\Omega}(\Omega_t).
\end{equation}
The body token provides a robust global context for the entire person. Since hand location and wrist orientation are strongly constrained by the upper body, shoulder, elbow, and global body pose, we use the body token as the central representation into which hand evidence is subsequently injected.

\subsubsection{Part-specific hand tokens}

In parallel, we encode left and right hand tokens using part-specific streams:
\begin{equation}
z_t^{h} = E_{K}^{h}(K_t^{h}) + E_F^h(F_t^{h}) + E_B^h(B_t^{h}),
\quad h \in \{lh, rh\}.
\end{equation}
The hand crop feature preserves fine visual details that are often lost in the body crop, while the hand keypoints and hand-box token provide explicit geometric evidence about finger layout, hand location, and scale. The left and right hand observations are encoded separately before fusion, preventing detailed hand information from being overwhelmed by the stronger body signal.

For keypoint observations, we use confidence-aware encoding. Visible joints are embedded from their 2D coordinates via a linear projection, while low-confidence joints are replaced by learned missing-joint embeddings. This allows the network to retain the identity of missing joints without trusting unreliable detector coordinates. The same principle is applied to body and hand keypoints, utilizing separate encoders for the body, left hand, and right hand.

Crucially, the hand streams are not independent hand predictors; they are local evidence encoders whose outputs are injected into the shared body-hand temporal representation.

\subsubsection{Residual body-hand fusion}

We inject hand evidence into the body-centric representation through a residual fusion design. The left and right hand tokens are concatenated and projected by an individual hand fusion MLP, denoted as $\phi_h$. The resulting hand evidence vector is added to the body token before temporal modeling:
\begin{equation}
z_t = z_t^b + \phi_h\left([z_t^{lh}, z_t^{rh}]\right).
\end{equation}
This design avoids two common failure modes: directly compressing whole-body observations through a single encoder can dilute fine hand details, while post-hoc stitching of hands and body predictions struggles with wrist-body consistency. 
By injecting hand evidence before temporal reasoning, the model jointly refines body pose, wrist orientation, and hand articulation.

Unlike early fusion by aggregating all observations, our residual fusion preserves the body representation as the main temporal carrier and treats hand evidence as a structured refinement. The residual formulation also improves robustness. When hand observations are unreliable, the body token provides a stable motion prior. With reliable hand evidence, the residual branch injects local hand details into the temporal model. This is particularly important for close-up videos, motion blur, and partial hand visibility.

The fused frame tokens are finally processed by a temporal motion encoder, which propagates information across frames and produces contextualized motion features $h_{1:T}$:
\begin{equation}
h_{1:T} = \mathcal{T}(z_{1:T}).
\end{equation}
We use relative temporal attention~\cite{su2024roformer} to support variable-length sequences and a local attention window for long clips.

\subsubsection{Motion decoding}

The contextualized temporal feature \(h_t\) is decoded into both human motion and camera-related outputs:
\begin{equation}
    \left(
    \theta_t^{x}, \beta_t, \Gamma_t^{GV}, \mathbf{v}_t^{root},
    \Gamma_t^{c}, \tau_t^{c}, \mathbf{p}_t
    \right)
    =
    H(h_t).
\end{equation}
To reduce foot sliding and unstable ground contact, we optionally apply a contact-aware trajectory refinement that optimizes root velocity with static-contact and jitter-suppression constraints.

\subsection{Close-up-aware training}

\subsubsection{Adaptive close-up view augmentation}

Real-world videos often contain close-up or upper-body framing, especially in livestream, vlog, and speech scenarios. 
To reduce the distribution gap between standard full-body training data and such videos, we synthesize close-up views from existing video datasets using 2D body keypoints.
Specifically, given the 2D body joints of a video clip, we adaptively crop the person based on an upper-body ratio:
\begin{equation}
    r_{\text{up}} = \frac{d_{\text{head-hip}}}{h_{\text{body}}},
\end{equation}
where $d_{\text{head-hip}}$ represents the head-to-hip joint distance and $h_{\text{body}}$ is the total visible height. For standing postures (higher $r_{\text{up}}$), the lower boundary is raised to prioritize the upper body, consistent with livestream and vlog-style framing. 
Conversely, for sitting postures (lower $r_{\text{up}}$), we preserve a larger lower-body region to prevent limb truncation. This pose-conditioned approach produces more realistic close-up views across diverse configurations than a fixed cropping strategy.

To further increase view diversity, we apply scale and translation perturbations to the crop, simulating different camera distances and off-center compositions. 
The perturbation is sampled at the clip level and shared by all frames, preserving temporal framing consistency and avoiding artificial crop jitter. 
Before resizing to the network input resolution, we apply a mild Gaussian blur when the crop is heavily downsampled to reduce aliasing artifacts.
During training, we mix the synthesized close-up views with the original full-body views. 

\subsubsection{Partial hand observation modeling}

Close-up view synthesis changes the visibility of hands: hands may remain fully visible, become partially truncated, shrink to a small region, or move outside the crop. 
We therefore recompute hand observations after view synthesis rather than reusing the original hand boxes.

For each hand, we define a visibility flag
\[
v_h =
\begin{cases}
1, & N_h \ge N_{\min} \ \text{and}\ s_h \ge s_{\min}, \\
0, & \text{otherwise},
\end{cases}
\]
where \(N_h\) denotes the number of reliable visible hand keypoints inside the synthesized crop, and \(s_h\) is the resulting hand-box scale.
A hand observation is treated as valid only when enough reliable keypoints remain visible, and the hand region is larger than a minimum hand box scale. 
Otherwise, the corresponding hand crop and hand-specific observations are masked.
This prevents invalid hand crops or hallucinated detector outputs from being treated as reliable evidence.

This design exposes the model to realistic hand visibility conditions during training.
When direct hand evidence is available, it can use local hand observations for detailed articulation. 
Otherwise, it learns to infer plausible hand motion from body context and temporal cues, matching real-world cases with occlusion, motion blur, and out-of-frame hands.

\subsection{Training objectives and curriculum}

We train the model with a compact SMPL-X reconstruction objective and introduce three task-specific designs for video whole-body recovery: hand-emphasized supervision, curriculum training, and visibility-aware hand supervision.

\subsubsection{Hand emphasis}

Applying a uniform loss across all SMPL-X joints is suboptimal for whole-body recovery. Because body motion is highly visible, it dominates the reconstruction loss, leaving hand articulation weakly supervised. This imbalance is especially detrimental in close-up videos, where minor finger errors cause noticeably unnatural hand motions.
We therefore re-weight hand pose and hand joint supervision, and additionally emphasize distal and fingertip joints:
\begin{equation}
\mathcal{L}_{tip}
=
\sum_{t}\sum_{j\in\mathcal{T}}
\|\mathbf{J}_{t,j}-\hat{\mathbf{J}}_{t,j}\|_1,
\end{equation}
where $\mathcal{T}$ denotes distal and fingertip joints. 
This fingertip-aware term directly constrains hand end-effectors, encouraging more accurate fine-grained articulation while preserving the full SMPL-X reconstruction objective.

\subsubsection{Curriculum training}

The training data contains heterogeneous supervision: large-scale body-centric videos provide broad motion coverage, while hand-rich whole-body data provides more detailed hand articulation. Training with a single uniform schedule can bias the model toward dominant body signals and under-emphasize fine-grained hand motion. We therefore adopt a two-stage curriculum.

\textbf{Stage I: General prior learning.} 
In the first stage, we train on the full data mixture with the standard SMPL-X reconstruction objective and temporal regularization. 
This stage aims to learn stable body motion, coarse hand pose, camera-space reconstruction, and gravity-aware global motion. 
The goal is to establish a broad whole-body motion prior before focusing on fine hand details.

\textbf{Stage II: Hand refinement.} 
In the second stage, we increase the sampling ratio of high-quality hand and whole-body data, and strengthen the supervision on hand joints, especially distal and fingertip joints.
Body and temporal supervision remain active during this stage to preserve whole-body consistency. 
This curriculum separates two learning objectives with different data requirements: diverse video data is used to learn robust body and global motion priors, while the later hand-focused stage refines the local articulation that is most important for visual quality.

\subsubsection{Visibility-aware supervision}

Real-world hand observations are frequently compromised by blur, occlusion, or truncation, causing 2D detectors to predict erroneous keypoints. Directly supervising these noisy inputs degrades temporal hand recovery.
We therefore apply the hand reprojection loss only to reliable 2D hand observations. Specifically, each detected hand keypoint is assigned a validity weight based on its detection confidence:
\begin{equation}
\mathcal{L}_{vis}
=
\frac{1}{\sum_{t,j} w_{t,j}}
\sum_{t,j} w_{t,j}
\left\|\Pi(\mathbf{J}_{t,j})-\hat{\mathbf{x}}_{t,j}\right\|_1 .
\end{equation}
Here, $\Pi$ denotes camera projection, $\hat{\mathbf{x}}_{t,j}$ is the detected 2D hand keypoint, and $w_{t,j}$ is the validity weight derived from the detector confidence. We additionally discard hand crops with too few reliable keypoints or extremely small box sizes.
Masking unreliable observations prevents noisy data from degrading training, allowing the temporal model to infer plausible hand motion from body context and adjacent valid frames.

\begin{table*}
    \caption{Quantitative comparison of whole-body reconstruction on ARCTIC~\cite{fan2023arctic} and UBody~\cite{lin2023osx}. 
    The best performance is in boldface. }
    \vspace{-8pt}
    \resizebox{1.0\linewidth}{!}
    {
    \begin{tabular}{l|cc|cc|cc|cc|cc|cc}
    \toprule
    \multirow{3}{*}{Method}  
    & \multicolumn{6}{c|}{\cellcolor{gray!15}\textbf{ARCTIC~\cite{fan2023arctic}}} 
    & \multicolumn{6}{c}{\cellcolor{gray!15}\textbf{UBody~\cite{lin2023osx}}} \\
    \cmidrule(lr){2-7} \cmidrule(lr){8-13}
    & \multicolumn{2}{c|}{PA-PVE $\downarrow$} 
    & \multicolumn{2}{c|}{PVE $\downarrow$} 
    & \multicolumn{2}{c|}{PA-MPJPE $\downarrow$} 
    & \multicolumn{2}{c|}{PA-PVE $\downarrow$} 
    & \multicolumn{2}{c|}{PVE $\downarrow$} 
    & \multicolumn{2}{c}{PA-MPJPE $\downarrow$} \\
    \cmidrule(lr){2-3} \cmidrule(lr){4-5} \cmidrule(lr){6-7}
    \cmidrule(lr){8-9} \cmidrule(lr){10-11} \cmidrule(lr){12-13}
    & All & Hands 
    & All & Hands 
    & All & Hands
    & All & Hands 
    & All & Hands 
    & All & Hands \\
    \midrule

    PIXIE~\cite{feng2021pixie} 
    & -- & -- & -- & -- & -- & -- 
    & 61.7 & 12.2 & 168.4 & 55.6 & 66.8 & 12.3 \\
    
    Hand4Whole~\cite{moon2022h4w} 
    & 63.4 & 18.1 & 136.8 & 54.8 & -- & --
    & 44.8 & 8.9 & 104.1 & 45.7 & 45.5 & \underline{9.0} \\

    OSX~\cite{lin2023osx} 
    & 33.0 & 18.8 & 58.4 & 39.4 & -- & --
    & 42.4 & 10.8 & 92.4 & 47.7 & 42.9 & 11.0 \\

    PyMAF-X~\cite{zhang2023pymafx} 
    & 49.4 & 14.3 & 70.5 & 45.1 & 56.7 & 15.6
    & -- & -- & -- & -- & -- & -- \\

    SMPLer-X-B~\cite{cai2024smplerx}
    & 54.0 & 17.9 & 85.2 & 53.4 & 55.1 & 18.4
    & 35.5 & 11.0 & 65.3 & 46.9 & 37.2 & 10.9 \\

    SMPLer-X-L~\cite{cai2024smplerx}
    & 46.9 & 18.1 & 76.9 & 50.8 & \underline{47.3} & 17.8
    & 33.2 & 10.6 & 61.5 & 43.3 & \underline{36.6} & 10.5 \\

    AiOS~\cite{sun2024aios} 
    & 30.2 & 19.2 & 47.1 & 38.3 & -- & --
    & 32.5 & \textbf{7.3} & \textbf{58.6} & 39.0 & -- & -- \\

    SMPLest-X~\cite{wanqiyin2026smplestx} 
    & \textbf{26.3} & \underline{11.9} & \textbf{43.5} & \underline{28.8} & -- & --
    & \underline{31.6} & 9.5 & \underline{59.8} & \underline{37.7} & -- & -- \\

    GVHMR~\cite{shen2024gvhmr}+HaMeR~\cite{pavlakos2024hamer}
    & 41.0 & 13.9 & 71.0 & 54.6 & 55.1 & \underline{11.7} 
    & -- & -- & -- & -- & -- & -- \\

    \textbf{Ours} 
    & \underline{28.1} & \textbf{8.5} & \underline{44.7} & \textbf{24.8} & \textbf{29.8} & \textbf{8.1}
    & \textbf{31.3} & \underline{7.4} & 61.0 & \textbf{22.2} & \textbf{34.6} & \textbf{4.8} \\
    \bottomrule
    \end{tabular}
    }
    \label{tab:whole_body}
\vspace{-5pt}
\end{table*}

\section{Experiments}

\subsection{Experimental setup}

\textbf{Datasets.}
For evaluation, we consider two categories of benchmarks. We first evaluate whole-body results on ARCTIC~\cite{fan2023arctic} and UBody~\cite{lin2023osx} datasets. ARCTIC is a controlled multi-view hand-object interaction dataset with accurate SMPL-X annotations, making it well-suited for evaluating articulated hand reconstruction and body-hand coordination. 
UBody contains diverse in-the-wild upper-body videos, including vlogss, video conferences, and sign language videos, in which severe truncation and expressive hand gestures pose significant challenges for whole-body recovery. 
We further report body-only reconstruction results on standard video HMR benchmarks, including 3DPW~\cite{3dpw} and EMDB~\cite{kaufmann2023emdb}.

\textbf{Metrics.}
For SMPL-X whole-body datasets, we report metrics for both the whole body and the hands, including PA-PVE, PVE, and PA-MPJPE. PA-PVE measures the Procrustes-aligned vertex error, PVE measures the raw vertex error, and PA-MPJPE measures the Procrustes-aligned joint error. These reconstruction errors are reported in millimeters ($\mathit{mm}$). 
For body-only video HMR benchmarks, we evaluate their body reconstruction errors.

To further evaluate temporal motion quality, we report MPJVE, Accel, and Jitter, which correspond to the errors in the first-, second-, and third-order temporal derivatives of the 3D joints, measured in $\mathit{mm}/\mathit{s}$, $\mathit{m}/\mathit{s}^2$, and $\mathit{m}/\mathit{s}^3$, respectively. MPJVE and Accel measure velocity and acceleration deviations from the ground truth, and Jitter quantifies the temporal instability of the motion. 
These metrics complement static frame-wise evaluations by capturing motion dynamics, acceleration consistency, and frame-to-frame smoothness.

\textbf{Implementation details.}
We use HMR2.0~\cite{goel2023hmr2.0} and HaMeR~\cite{pavlakos2024hamer} as the frozen body and hand image feature extractors, respectively. We use ViTPose~\cite{xu2023ViTPose++} and RTMW~\cite{jiang2024rtmw,jiang2023rtmpose} to extract 2D body and hand keypoints. 
During training, all external detectors and feature extractors are kept frozen, and only our temporal reconstruction network is optimized.
Hand keypoints with confidence lower than 0.55 are masked out. For close-up augmented samples, we recompute hand boxes after cropping and discard a hand observation if fewer than 5 valid hand keypoints remain visible or if the hand box scale is smaller than 20 pixels. 
The temporal window size is 120.
We train the model in two stages. Stage I uses the full mixed training set for 207k steps. Stage II runs for another 88k steps, with increased sampling of high-quality SMPL-X and hand-centric data.

\begin{table}[h]
\centering
\caption{Quantitative comparison of body-only reconstruction on 3DPW~\cite{3dpw} and EMDB~\cite{kaufmann2023emdb}.}
\vspace{-5pt}
\resizebox{1.0\linewidth}{!}{
    \begin{tabular}{lcccc}
    \toprule
    Method & PA-MPJPE $\downarrow$ & MPJPE $\downarrow$ & PVE $\downarrow$ & Accel $\downarrow$ \\
    \midrule
    \rowcolor{gray!15}
    \multicolumn{5}{l}{\textbf{3DPW~\cite{3dpw}}} \\
    \cmidrule(lr){1-5}
    
    GVHMR~\cite{shen2024gvhmr} & 37.0 & 56.6 & 68.7 & \underline{5.2} \\
    GENMO~\cite{genmo2025} & \textbf{34.6} & 53.9 & 65.8 & \underline{5.2} \\
    PromptHMR-Vid~\cite{wang2025prompthmr} & 35.5 & 56.9 & 67.3 & -- \\
    \cmidrule(lr){1-5}
    OSX~\cite{lin2023osx} & 60.6 & 86.2 & -- & -- \\
    SMPLer-X-B~\cite{cai2024smplerx} & 53.4 & 80.3 & 90.3 & -- \\
    Multi-HMR~\cite{baradel2024multihmr} & 41.7 & 61.4 & 75.9 & -- \\
    SMPLest-X~\cite{wanqiyin2026smplestx} & 43.2 & 70.5 & -- & -- \\
    \textbf{Ours} & \underline{35.2} & \textbf{53.6} & \textbf{65.1} & \textbf{4.9} \\
    
    \midrule
    \rowcolor{gray!15}
    \multicolumn{5}{l}{\textbf{EMDB~\cite{kaufmann2023emdb}}} \\
    \cmidrule(lr){1-5}
    
    GVHMR~\cite{shen2024gvhmr} & 44.5 & 74.2 & 85.9 & 3.9 \\
    GENMO~\cite{genmo2025} & 42.5 & 73.0 & 84.8 & \underline{3.8} \\
    PromptHMR-Vid~\cite{wang2025prompthmr} & \textbf{40.1} & 68.1 & 79.2 & -- \\
    DUOMO~\cite{wang2026duomo} & \underline{41.8} & \underline{67.1} & \underline{78.2} & -- \\
    \cmidrule(lr){1-5}
    Multi-HMR~\cite{baradel2024multihmr} & 50.1 & 81.6 & 95.7 & -- \\
    \textbf{Ours} & \textbf{40.1} & \textbf{64.3} & \textbf{75.4} & \textbf{3.5} \\
    
    \bottomrule
\end{tabular}
}
\label{tab:body_only}
\vspace{-5pt}
\end{table}

\subsection{Whole-body reconstruction}
\subsubsection{Quantitative results}
Table~\ref{tab:whole_body} reports whole-body reconstruction results on ARCTIC and UBody. 
ARCTIC emphasizes articulated hand motion in hand-object interaction, and UBody focuses more on in-the-wild upper-body videos with truncation and expressive gestures. 
Consequently, these benchmarks serve to evaluate the hand reconstruction fidelity and the robustness of the proposed method in close-up whole-body recovery scenarios, respectively.
On the ARCTIC dataset, the proposed method outperforms existing approaches across all reported hand metrics. Specifically, compared to the previous state-of-the-art, SMPLest-X, our approach achieves a $28.6\%$ reduction in hand PA-PVE and a $13.9\%$ reduction in hand PVE. Similarly, our method attains the lowest hand PVE on UBody. These quantitative improvements demonstrate that temporal body-hand modeling significantly enhances reconstruction accuracy in both controlled interaction settings and challenging, unconstrained video environments.

Notably, our method uses a much smaller model than recent whole-body foundation models while achieving comparable or even better performance.
SMPLer-X-B, SMPLer-X-L, and SMPLest-X contain $103$M, $327$M, and $687$M trainable parameters, respectively, while our model has only about $46$M parameters and is trained with much less data.
Despite this increased efficiency, the method achieves competitive whole-body accuracy and exhibits markedly superior performance in hand reconstruction.

\subsubsection{Qualitative results}
\begin{figure*}[h]
  \centering
  \includegraphics[width=1.0\linewidth]{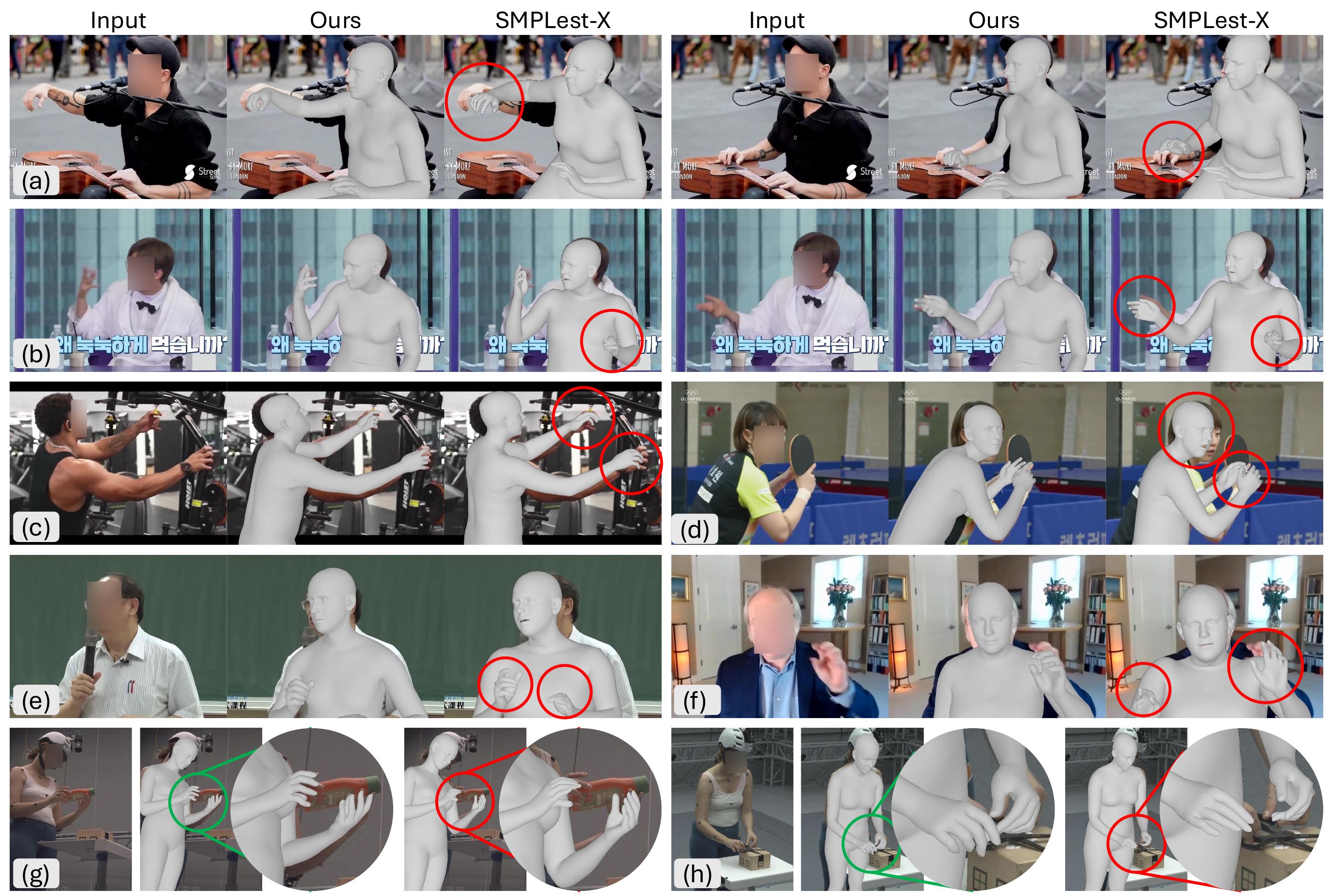}
  \vspace{-15pt}
  \caption{Qualitative results on UBody~\cite{lin2023osx} (a$\sim$f) and ARCTIC~\cite{fan2023arctic} (g$\sim$h). Compared with the image-based SMPLest-X~\cite{wanqiyin2026smplestx} baseline, which often fails in close-up and partial-body scenes, especially for hand pose estimation, our method produces more stable and plausible body-hand reconstructions across diverse half-body and hand-centric scenarios. Please zoom in for better visualization.}
  \label{fig:benchmark_vis}
\end{figure*}

Fig.~\ref{fig:benchmark_vis} presents qualitative comparisons on UBody and ARCTIC. Our method shows clear advantages over the image-based baseline SMPLest-X in half-body and hand-centric scenarios. In cases where a hand is outside the visible image region, SMPLest-X still tends to hallucinate implausible hand poses, as shown in cases (b), (e), and (f). By contrast, our method produces more reasonable predictions and avoids such spurious hand reconstructions. Our approach is also more robust under challenging conditions, such as hand occlusion (cases (d), (g), and (h)) and motion blur (cases (b) and (f)), where it still recovers plausible hand poses and coherent body-hand articulation. We attribute these improvements to our joint temporal modeling of body and hands, which allows the model to leverage temporal context and body motion cues to stabilize hand reconstruction. In contrast, the frame-wise image-based baseline lacks temporal reasoning and therefore yields noticeably less reliable hand predictions.

\subsection{Body-only evaluation}

Table~\ref{tab:body_only} reports the comparison on standard body-only video HMR benchmarks, including 3DPW and EMDB. 
We compare our method with both video-based body-only methods and recent whole-body models.
Even though our model is designed for whole-body SMPL-X reconstruction, it achieves state-of-the-art body reconstruction performance on both datasets, showing that our method preserves strong body pose accuracy. 
Compared with the strongest prior method DUOMO~\cite{wang2026duomo}, our method reduces MPJPE by $4.2\%$ on the EMDB benchmark. 
These results show that our method achieves state-of-the-art body-only reconstruction while supporting temporally coherent whole-body and hand recovery.

\begin{table}[t]
\centering
\caption{Temporal stability comparison on ARCTIC~\cite{fan2023arctic}. Lower values indicate smoother and more accurate temporal motion.}
\vspace{-5pt}
\resizebox{1.0\linewidth}{!}{
    \begin{tabular}{l|ccc|ccc}
    \toprule
    \multirow{2}{*}{Method}
    & \multicolumn{3}{c|}{\textbf{All}}
    & \multicolumn{3}{c}{\textbf{Hands}} \\
    \cmidrule(lr){2-4} \cmidrule(lr){5-7}
    & Jitter $\downarrow$
    & Accel $\downarrow$
    & MPJVE $\downarrow$
    & Jitter $\downarrow$
    & Accel $\downarrow$
    & MPJVE $\downarrow$ \\
    \midrule
    
    SMPLer-X-B~\cite{cai2024smplerx} & 127.5 & 23.8 & 482.6 & 52.1 & 9.9 & 219.9 \\
    SMPLest-X~\cite{wanqiyin2026smplestx} & 95.1 & 17.8 & 364.6 & 55.4 & 10.4 & 228.5 \\
    GVHMR~\cite{shen2024gvhmr}+HaMeR~\cite{pavlakos2024hamer} & 6.0 & 1.8 & 109.0 & 7.4 & 1.6 & 100.7 \\
    \textbf{Ours} & 5.4 & 1.5 & 81.3 & 4.2 & 1.5 & 67.7 \\
    
    \bottomrule
    \end{tabular}
    }
\label{tab:temporal}
\vspace{-5pt}
\end{table}

\subsection{Temporal motion quality}

Table~\ref{tab:temporal} evaluates temporal consistency on ARCTIC using Jitter, Accel, and MPJVE. 
These metrics measure complementary aspects of motion stability: Jitter reflects high-frequency temporal fluctuation, Accel measures acceleration-level motion error, and MPJVE evaluates velocity consistency with the ground-truth motion. 

Image-based whole-body methods, such as SMPLer-X and SMPLest-X, suffer from severe temporal inconsistency because they reconstruct each frame independently. 
In contrast, GVHMR+HaMeR provides a much stronger baseline by combining a video-based body model with an image-based hand estimator, leading to significantly lower temporal errors than purely image-based methods. 
However, since the hands are still estimated independently by HaMeR and then stitched to the body, this pipeline cannot fully ensure temporally coherent body-hand motion. 
In contrast, our method predicts body and hands jointly in a temporal framework, leading to more stable whole-body motion and significantly smoother hand trajectories.
Compared with GVHMR+HaMeR, our method reduces hand Jitter by $43.2\%$ and hand MPJVE by $32.8\%$, demonstrating the benefit of joint temporal body-hand modeling. 

\begin{table}[t]
\centering
\caption{
Ablation study on ARCTIC~\cite{fan2023arctic}. 
We report whole-body (All) PVE to measure overall whole-body reconstruction quality, and Hand PA-PVE, Hand PVE, and Hand Jitter to evaluate hand accuracy and temporal stability.
}
\vspace{-5pt}
\resizebox{1.0\linewidth}{!}{
\begin{tabular}{l|cccc}
\toprule
\multirow{2}{*}{Variant}
& \multicolumn{4}{c}{ARCTIC~\cite{fan2023arctic}} \\
\cmidrule(lr){2-5}
& All PVE $\downarrow$
& Hand PA-PVE $\downarrow$
& Hand PVE $\downarrow$
& Hand Jitter $\downarrow$ \\
\midrule

\textbf{Ours}
& 44.7 & 8.5 & 24.8 & 4.2 \\

\textit{(1) w/o Res fusion}
& 49.2 & 9.4 & 28.2 & 6.6 \\

\textit{(2) w/o Hand obs.}
& 55.2 & 15.8 & 44.8 & 10.1 \\

\textit{(3) w/o Close-up aug.}
& 48.4 & 9.2 & 31.1 & 5.9 \\

\textit{(4) w/o Two-stage}      
& 49.9 & 10.5 & 31.7 & 4.8 \\

\textit{(5) w/o $\mathcal{L}_{vis}$}
& 46.1 & 9.3 & 27.3 & 5.7 \\

\bottomrule
\end{tabular}
}
\label{tab:ablation}
\vspace{-10pt}
\end{table}

\subsection{Ablation study}

As shown in Table~\ref{tab:ablation}, we conduct ablation studies to validate the core architectural and training designs of our framework. 
We evaluate five variants on ARCTIC, focusing on both overall reconstruction quality and hand-centric performance. 
Specifically, (1) replaces our residual body-hand fusion with early fusion of all features, (2) removes part-specific hand observations and relies mainly on body-level cues, (3) disables close-up view augmentation during training, (4) removes the second-stage hand-focused training, and (5) removes the visibility-aware hand supervision term $\mathcal{L}_{vis}$.

The comparison shows that each design contributes to hand reconstruction quality and temporal stability. 
Variant (1) verifies the importance of structured body-hand fusion: directly compressing all body and hand features through early fusion weakens the interaction between global body context and local hand evidence, leading to degraded hand accuracy and jitter. 
Variant (2) produces a clear drop in hand-related metrics, confirming that body-level observations alone are insufficient for fine-grained hand articulation and that part-specific hand cues are essential. 
Variant (3) mainly affects robustness, where upper-body framing, truncation, and hand entry/exit are common, demonstrating the benefit of close-up augmentation for in-the-wild videos. 
Variant (4) shows that the second-stage training further improves hand reconstruction by emphasizing high-quality SMPL-X and hand-centric data. 
Finally, variant (5) highlights the importance of visibility-aware supervision: without masking invisible or unreliable hand regions, noisy supervision from occluded or out-of-frame hands harms both hand accuracy and temporal stability. 
Overall, the full model consistently achieves the best balance between whole-body accuracy, hand reconstruction quality, and temporal smoothness.

\subsection{Qualitative results on real-world videos}

\begin{figure*}[b]
  \centering
  \includegraphics[width=0.6\linewidth]{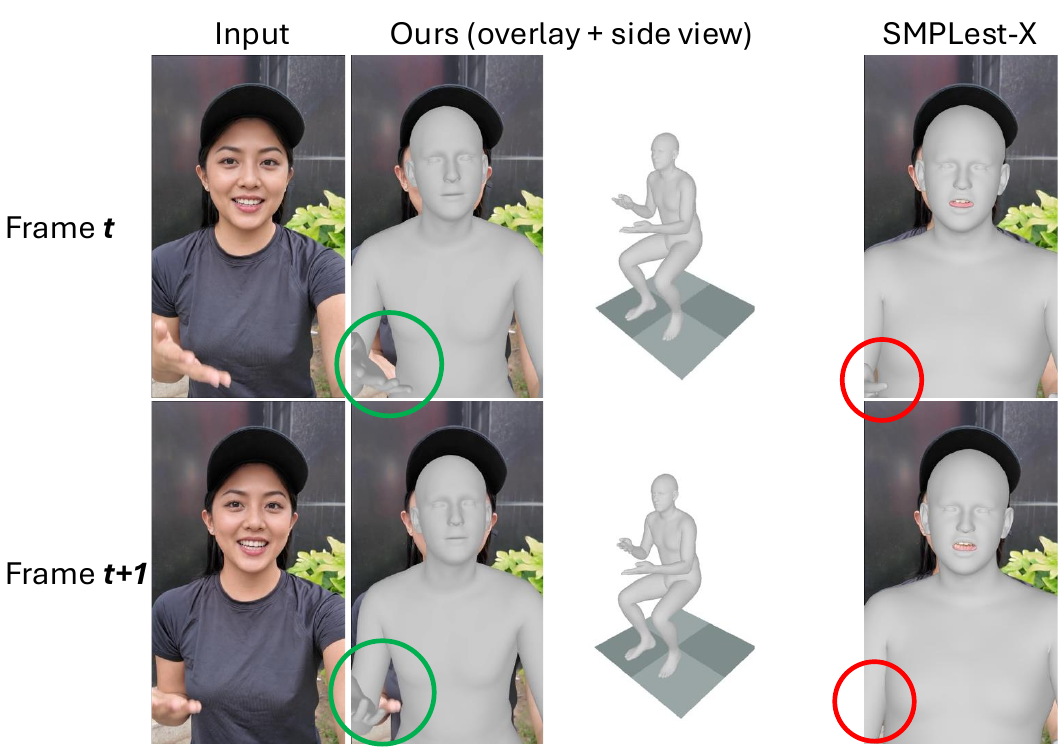}
  \vspace{-5pt}
  \caption{Qualitative comparison on a close-up internet video with frequent hand truncation. Since the right hand repeatedly enters and leaves the frame, SMPLest-X produces inconsistent hand poses across neighboring frames, resulting in visible jitter. Our method recovers more natural and temporally coherent hand motion, while preserving plausible body poses.}
  \label{fig:partial_vis}
\end{figure*}

\begin{figure*}[h]
  \centering
  \includegraphics[width=1.0\linewidth]{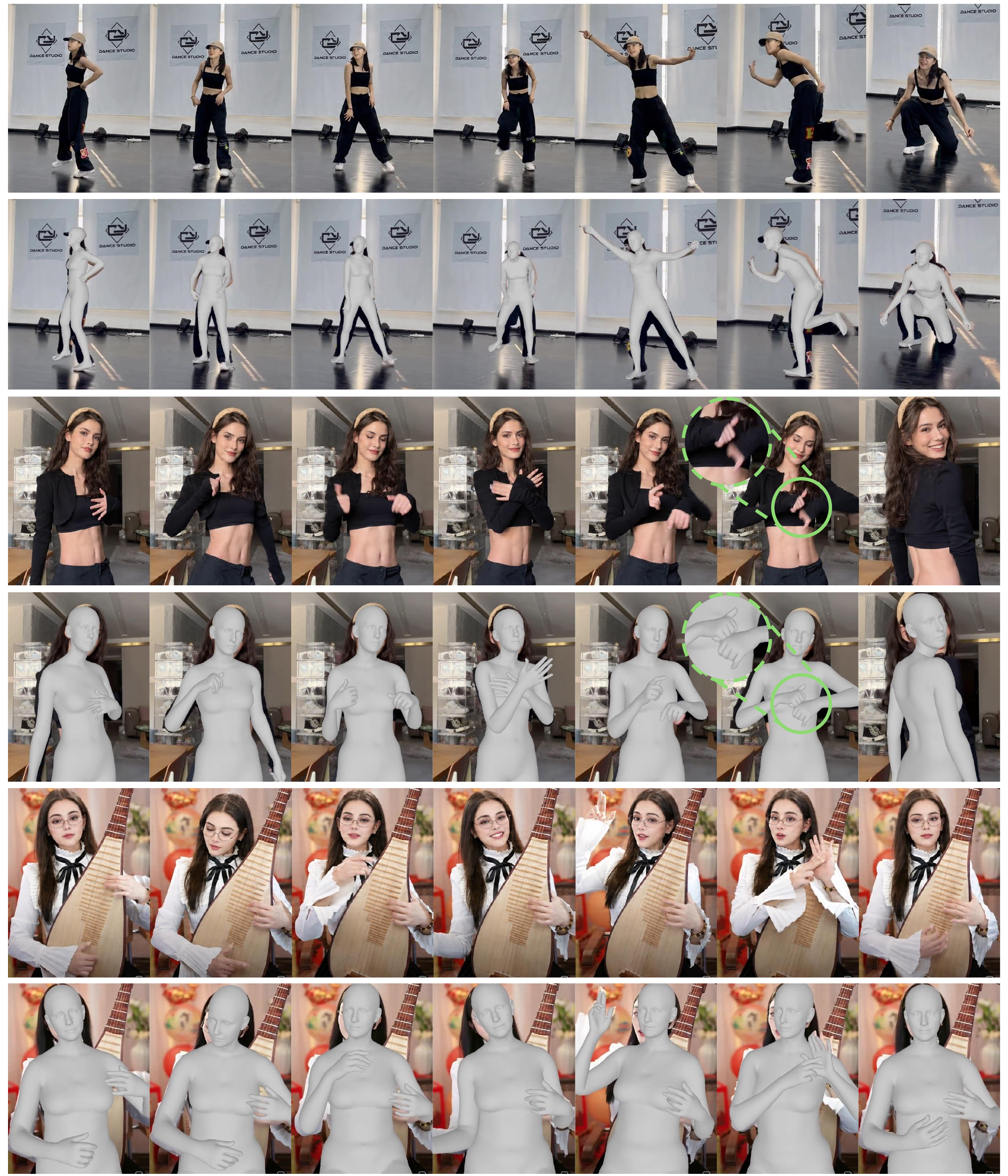}
  \vspace{-10pt}
  \caption{Qualitative results of our DanceHMR on challenging in-the-wild internet videos. For each example, we show input video frames and our reconstructed SMPL-X overlay. 
  DanceHMR recovers high-quality body and hand motion in fast dance, close-up, and livestream scenarios, maintaining stable hands and coherent whole-body dynamics. Please zoom in for better visualization.}
  \label{fig:itw_vis}
\end{figure*}

Fig.~\ref{fig:itw_vis} shows qualitative results on challenging in-the-wild internet videos. Our method produces stable and coherent whole-body motion across diverse scenarios, including agile hand interactions, severe motion blur, and complex body movements. 
Fig.~\ref{fig:partial_vis} shows a challenging close-up video with frequent hand truncation. Compared with the image-based SMPLest-X baseline, our method produces significantly more stable and temporally coherent hand motion under partial visibility. More qualitative results can be found in the supplementary video.
\section{Conclusion}
In this paper, we present DanceHMR, a temporally coherent body and hand human mesh recovery framework for monocular videos. Our method jointly models body motion and hand articulation through residual body-hand fusion within a unified temporal architecture, and improves robustness with close-up-aware augmentation. Experiments demonstrate improved hand reconstruction, competitive body accuracy, and stable SMPL-X motion in challenging real-world videos.


\bibliographystyle{plainnat}
\bibliography{main}

\begin{thebibliography}{37}
\providecommand{\natexlab}[1]{#1}
\providecommand{\url}[1]{\texttt{#1}}
\expandafter\ifx\csname urlstyle\endcsname\relax
  \providecommand{\doi}[1]{doi: #1}\else
  \providecommand{\doi}{doi: \begingroup \urlstyle{rm}\Url}\fi

\bibitem[Baradel et~al.(2024)Baradel, Armando, Galaaoui, Br{\'e}gier, Weinzaepfel, Rogez, and Lucas]{baradel2024multihmr}
Fabien Baradel, Matthieu Armando, Salma Galaaoui, Romain Br{\'e}gier, Philippe Weinzaepfel, Gr{\'e}gory Rogez, and Thomas Lucas.
\newblock Multi-hmr: Multi-person whole-body human mesh recovery in a single shot.
\newblock In \emph{European Conference on Computer Vision}, pages 202--218. Springer, 2024.

\bibitem[Cai et~al.(2024)Cai, Yin, Zeng, Wei, Sun, Yanjun, Pang, Mei, Zhang, Zhang, et~al.]{cai2024smplerx}
Zhongang Cai, Wanqi Yin, Ailing Zeng, Chen Wei, Qingping Sun, Wang Yanjun, Hui~En Pang, Haiyi Mei, Mingyuan Zhang, Lei Zhang, et~al.
\newblock Smpler-x: Scaling up expressive human pose and shape estimation.
\newblock \emph{Advances in Neural Information Processing Systems}, 36, 2024.

\bibitem[Chen et~al.(2025)Chen, Chen, Xue, Chen, Xiu, and Gerard]{chen2025human3r}
Yue Chen, Xingyu Chen, Yuxuan Xue, Anpei Chen, Yuliang Xiu, and Pons-Moll Gerard.
\newblock Human3r: Everyone everywhere all at once.
\newblock \emph{arXiv preprint arXiv:2510.06219}, 2025.

\bibitem[Choi et~al.(2021)Choi, Moon, Chang, and Lee]{choi2021tcmr}
Hongsuk Choi, Gyeongsik Moon, Ju~Yong Chang, and Kyoung~Mu Lee.
\newblock Beyond static features for temporally consistent 3d human pose and shape from a video.
\newblock In \emph{Proceedings of the IEEE/CVF conference on computer vision and pattern recognition}, pages 1964--1973, 2021.

\bibitem[Choutas et~al.(2020)Choutas, Pavlakos, Bolkart, Tzionas, and Black]{choutas2020expose}
Vasileios Choutas, Georgios Pavlakos, Timo Bolkart, Dimitrios Tzionas, and Michael~J Black.
\newblock Monocular expressive body regression through body-driven attention.
\newblock In \emph{Computer Vision--ECCV 2020: 16th European Conference, Glasgow, UK, August 23--28, 2020, Proceedings, Part X 16}, pages 20--40. Springer, 2020.

\bibitem[Fan et~al.(2023)Fan, Taheri, Tzionas, Kocabas, Kaufmann, Black, and Hilliges]{fan2023arctic}
Zicong Fan, Omid Taheri, Dimitrios Tzionas, Muhammed Kocabas, Manuel Kaufmann, Michael~J Black, and Otmar Hilliges.
\newblock Arctic: A dataset for dexterous bimanual hand-object manipulation.
\newblock In \emph{Proceedings of the IEEE/CVF Conference on Computer Vision and Pattern Recognition}, pages 12943--12954, 2023.

\bibitem[Feng et~al.(2021)Feng, Choutas, Bolkart, Tzionas, and Black]{feng2021pixie}
Yao Feng, Vasileios Choutas, Timo Bolkart, Dimitrios Tzionas, and Michael~J Black.
\newblock Collaborative regression of expressive bodies using moderation.
\newblock In \emph{2021 International Conference on 3D Vision (3DV)}, pages 792--804. IEEE, 2021.

\bibitem[Goel et~al.(2023)Goel, Pavlakos, Rajasegaran, Kanazawa, and Malik]{goel2023hmr2.0}
Shubham Goel, Georgios Pavlakos, Jathushan Rajasegaran, Angjoo Kanazawa, and Jitendra Malik.
\newblock Humans in 4d: Reconstructing and tracking humans with transformers.
\newblock In \emph{Proceedings of the IEEE/CVF International Conference on Computer Vision}, pages 14783--14794, 2023.

\bibitem[Jiang et~al.(2023)Jiang, Lu, Zhang, Ma, Han, Lyu, Li, and Chen]{jiang2023rtmpose}
Tao Jiang, Peng Lu, Li~Zhang, Ningsheng Ma, Rui Han, Chengqi Lyu, Yining Li, and Kai Chen.
\newblock Rtmpose: Real-time multi-person pose estimation based on mmpose, 2023.
\newblock URL \url{https://arxiv.org/abs/2303.07399}.

\bibitem[Jiang et~al.(2024)Jiang, Xie, and Li]{jiang2024rtmw}
Tao Jiang, Xinchen Xie, and Yining Li.
\newblock Rtmw: Real-time multi-person 2d and 3d whole-body pose estimation.
\newblock \emph{arXiv preprint arXiv:2407.08634}, 2024.

\bibitem[Kaufmann et~al.(2023)Kaufmann, Song, Guo, Shen, Jiang, Tang, Z{\'a}rate, and Hilliges]{kaufmann2023emdb}
Manuel Kaufmann, Jie Song, Chen Guo, Kaiyue Shen, Tianjian Jiang, Chengcheng Tang, Juan~Jos{\'e} Z{\'a}rate, and Otmar Hilliges.
\newblock {EMDB}: The {E}lectromagnetic {D}atabase of {G}lobal 3{D} {H}uman {P}ose and {S}hape in the {W}ild.
\newblock In \emph{International Conference on Computer Vision (ICCV)}, 2023.

\bibitem[Kocabas et~al.(2020)Kocabas, Athanasiou, and Black]{kocabas2020vibe}
Muhammed Kocabas, Nikos Athanasiou, and Michael~J Black.
\newblock Vibe: Video inference for human body pose and shape estimation.
\newblock In \emph{Proceedings of the IEEE/CVF conference on computer vision and pattern recognition}, pages 5253--5263, 2020.

\bibitem[Li et~al.(2025)Li, Cao, Zhang, Rempe, Kautz, Iqbal, and Yuan]{genmo2025}
Jiefeng Li, Jinkun Cao, Haotian Zhang, Davis Rempe, Jan Kautz, Umar Iqbal, and Ye~Yuan.
\newblock Genmo: A generalist model for human motion.
\newblock In \emph{Proceedings of the IEEE/CVF International Conference on Computer Vision (ICCV)}, 2025.

\bibitem[Lin et~al.(2023)Lin, Zeng, Wang, Zhang, and Li]{lin2023osx}
Jing Lin, Ailing Zeng, Haoqian Wang, Lei Zhang, and Yu~Li.
\newblock One-stage 3d whole-body mesh recovery with component aware transformer.
\newblock In \emph{Proceedings of the IEEE/CVF Conference on Computer Vision and Pattern Recognition}, pages 21159--21168, 2023.

\bibitem[Moon et~al.(2022)Moon, Choi, and Lee]{moon2022h4w}
Gyeongsik Moon, Hongsuk Choi, and Kyoung~Mu Lee.
\newblock Accurate 3d hand pose estimation for whole-body 3d human mesh estimation.
\newblock In \emph{Proceedings of the IEEE/CVF Conference on Computer Vision and Pattern Recognition}, pages 2308--2317, 2022.

\bibitem[Pavlakos et~al.(2024)Pavlakos, Shan, Radosavovic, Kanazawa, Fouhey, and Malik]{pavlakos2024hamer}
Georgios Pavlakos, Dandan Shan, Ilija Radosavovic, Angjoo Kanazawa, David Fouhey, and Jitendra Malik.
\newblock Reconstructing hands in 3d with transformers.
\newblock In \emph{Proceedings of the IEEE/CVF Conference on Computer Vision and Pattern Recognition}, pages 9826--9836, 2024.

\bibitem[Potamias et~al.(2025)Potamias, Zhang, Deng, and Zafeiriou]{potamias2025wilor}
Rolandos~Alexandros Potamias, Jinglei Zhang, Jiankang Deng, and Stefanos Zafeiriou.
\newblock Wilor: End-to-end 3d hand localization and reconstruction in-the-wild.
\newblock In \emph{Proceedings of the Computer Vision and Pattern Recognition Conference}, pages 12242--12254, 2025.

\bibitem[Rempe et~al.(2021)Rempe, Birdal, Hertzmann, Yang, Sridhar, and Guibas]{rempe2021humor}
Davis Rempe, Tolga Birdal, Aaron Hertzmann, Jimei Yang, Srinath Sridhar, and Leonidas~J Guibas.
\newblock Humor: 3d human motion model for robust pose estimation.
\newblock In \emph{Proceedings of the IEEE/CVF international conference on computer vision}, pages 11488--11499, 2021.

\bibitem[Shen et~al.(2024{\natexlab{a}})Shen, Yin, Wang, Wei, Cai, Yang, and Lin]{hmradapter}
Wenhao Shen, Wanqi Yin, Hao Wang, Chen Wei, Zhongang Cai, Lei Yang, and Guosheng Lin.
\newblock Hmr-adapter: A lightweight adapter with dual-path cross augmentation for expressive human mesh recovery.
\newblock In \emph{Proceedings of the 32nd ACM International Conference on Multimedia}, pages 6093--6102, 2024{\natexlab{a}}.

\bibitem[Shen et~al.(2023)Shen, Yang, Wang, Ma, Zhou, and Yang]{shen2023glot}
Xiaolong Shen, Zongxin Yang, Xiaohan Wang, Jianxin Ma, Chang Zhou, and Yi~Yang.
\newblock Global-to-local modeling for video-based 3d human pose and shape estimation.
\newblock In \emph{Proceedings of the IEEE/CVF Conference on Computer Vision and Pattern Recognition}, pages 8887--8896, 2023.

\bibitem[Shen et~al.(2024{\natexlab{b}})Shen, Pi, Xia, Cen, Peng, Hu, Bao, Hu, and Zhou]{shen2024gvhmr}
Zehong Shen, Huaijin Pi, Yan Xia, Zhi Cen, Sida Peng, Zechen Hu, Hujun Bao, Ruizhen Hu, and Xiaowei Zhou.
\newblock World-grounded human motion recovery via gravity-view coordinates.
\newblock In \emph{SIGGRAPH Asia 2024 Conference Papers}, pages 1--11, 2024{\natexlab{b}}.

\bibitem[Shin et~al.(2024)Shin, Kim, Halilaj, and Black]{shin2024wham}
Soyong Shin, Juyong Kim, Eni Halilaj, and Michael~J Black.
\newblock Wham: Reconstructing world-grounded humans with accurate 3d motion.
\newblock In \emph{Proceedings of the IEEE/CVF Conference on Computer Vision and Pattern Recognition}, pages 2070--2080, 2024.

\bibitem[Su et~al.(2024)Su, Ahmed, Lu, Pan, Bo, and Liu]{su2024roformer}
Jianlin Su, Murtadha Ahmed, Yu~Lu, Shengfeng Pan, Wen Bo, and Yunfeng Liu.
\newblock Roformer: Enhanced transformer with rotary position embedding.
\newblock \emph{Neurocomputing}, 568:\penalty0 127063, 2024.

\bibitem[Sun et~al.(2024)Sun, Wang, Zeng, Yin, Wei, Wang, Mei, Leung, Liu, Yang, et~al.]{sun2024aios}
Qingping Sun, Yanjun Wang, Ailing Zeng, Wanqi Yin, Chen Wei, Wenjia Wang, Haiyi Mei, Chi-Sing Leung, Ziwei Liu, Lei Yang, et~al.
\newblock Aios: All-in-one-stage expressive human pose and shape estimation.
\newblock In \emph{Proceedings of the IEEE/CVF conference on computer vision and pattern recognition}, pages 1834--1843, 2024.

\bibitem[Sun et~al.(2023)Sun, Bao, Liu, Mei, and Black]{sun2023trace}
Yu~Sun, Qian Bao, Wu~Liu, Tao Mei, and Michael~J Black.
\newblock Trace: 5d temporal regression of avatars with dynamic cameras in 3d environments.
\newblock In \emph{Proceedings of the IEEE/CVF Conference on Computer Vision and Pattern Recognition}, pages 8856--8866, 2023.

\bibitem[Von~Marcard et~al.(2018)Von~Marcard, Henschel, Black, Rosenhahn, and Pons-Moll]{3dpw}
Timo Von~Marcard, Roberto Henschel, Michael~J Black, Bodo Rosenhahn, and Gerard Pons-Moll.
\newblock Recovering accurate 3d human pose in the wild using imus and a moving camera.
\newblock In \emph{Proceedings of the European conference on computer vision (ECCV)}, pages 601--617, 2018.

\bibitem[Wang et~al.(2024)Wang, Wang, Liu, and Daniilidis]{wang2024tram}
Yufu Wang, Ziyun Wang, Lingjie Liu, and Kostas Daniilidis.
\newblock Tram: Global trajectory and motion of 3d humans from in-the-wild videos.
\newblock In \emph{European Conference on Computer Vision}, pages 467--487. Springer, 2024.

\bibitem[Wang et~al.(2025)Wang, Sun, Patel, Daniilidis, Black, and Kocabas]{wang2025prompthmr}
Yufu Wang, Yu~Sun, Priyanka Patel, Kostas Daniilidis, Michael~J Black, and Muhammed Kocabas.
\newblock Prompthmr: Promptable human mesh recovery.
\newblock In \emph{Proceedings of the computer vision and pattern recognition conference}, pages 1148--1159, 2025.

\bibitem[Wang et~al.(2026)Wang, Ng, Shin, Khirodkar, Dong, Su, Park, Kitani, Richard, Prada, and Zollhofer]{wang2026duomo}
Yufu Wang, Evonne Ng, Soyong Shin, Rawal Khirodkar, Yuan Dong, Zhaoen Su, Jinhyung Park, Kris Kitani, Alexander Richard, Fabian Prada, and Michael Zollhofer.
\newblock Duomo: Dual motion diffusion for world-space human reconstruction.
\newblock \emph{arXiv preprint arXiv:2603.03265}, 2026.

\bibitem[Wei et~al.(2022)Wei, Lin, Liu, and Liao]{wei2022mpsnet}
Wen-Li Wei, Jen-Chun Lin, Tyng-Luh Liu, and Hong-Yuan~Mark Liao.
\newblock Capturing humans in motion: Temporal-attentive 3d human pose and shape estimation from monocular video.
\newblock In \emph{Proceedings of the IEEE/CVF Conference on Computer Vision and Pattern Recognition}, pages 13211--13220, 2022.

\bibitem[Xu et~al.(2024)Xu, Zhang, Zhang, and Tao]{xu2023ViTPose++}
Yufei Xu, Jing Zhang, Qiming Zhang, and Dacheng Tao.
\newblock Vitpose++: Vision transformer foundation model for generic body pose estimation.
\newblock \emph{IEEE Transactions on Pattern Analysis and Machine Intelligence}, 46:\penalty0 1212--1230, 2024.
\newblock \doi{10.1109/TPAMI.2023.3330016}.

\bibitem[Ye et~al.(2023)Ye, Pavlakos, Malik, and Kanazawa]{ye2023slahmr}
Vickie Ye, Georgios Pavlakos, Jitendra Malik, and Angjoo Kanazawa.
\newblock Decoupling human and camera motion from videos in the wild.
\newblock In \emph{Proceedings of the IEEE/CVF conference on computer vision and pattern recognition}, pages 21222--21232, 2023.

\bibitem[Yin et~al.(2024)Yin, Cai, Wang, Wang, Wei, Mei, Xiao, Yang, Sun, Yamashita, et~al.]{yin2024whac}
Wanqi Yin, Zhongang Cai, Ruisi Wang, Fanzhou Wang, Chen Wei, Haiyi Mei, Weiye Xiao, Zhitao Yang, Qingping Sun, Atsushi Yamashita, et~al.
\newblock Whac: World-grounded humans and cameras.
\newblock In \emph{European Conference on Computer Vision}, pages 20--37. Springer, 2024.

\bibitem[Yin et~al.(2026)Yin, Cai, Wang, Zeng, Wei, Sun, Mei, Wang, Pang, Zhang, Zhang, Loy, Yamashita, Yang, and Liu]{wanqiyin2026smplestx}
Wanqi Yin, Zhongang Cai, Ruisi Wang, Ailing Zeng, Chen Wei, Qingping Sun, Haiyi Mei, Yanjun Wang, Hui~En Pang, Mingyuan Zhang, Lei Zhang, Chen~Change Loy, Atsushi Yamashita, Lei Yang, and Ziwei Liu.
\newblock { SMPLest-X: Ultimate Scaling for Expressive Human Pose and Shape Estimation }.
\newblock \emph{IEEE Transactions on Pattern Analysis \& Machine Intelligence}, 48\penalty0 (02):\penalty0 1778--1794, February 2026.
\newblock ISSN 1939-3539.
\newblock \doi{10.1109/TPAMI.2025.3618174}.

\bibitem[Yuan et~al.(2022)Yuan, Iqbal, Molchanov, Kitani, and Kautz]{yuan2022glamr}
Ye~Yuan, Umar Iqbal, Pavlo Molchanov, Kris Kitani, and Jan Kautz.
\newblock Glamr: Global occlusion-aware human mesh recovery with dynamic cameras.
\newblock In \emph{Proceedings of the IEEE/CVF conference on computer vision and pattern recognition}, pages 11038--11049, 2022.

\bibitem[Zhang et~al.(2023)Zhang, Tian, Zhang, Li, An, Sun, and Liu]{zhang2023pymafx}
Hongwen Zhang, Yating Tian, Yuxiang Zhang, Mengcheng Li, Liang An, Zhenan Sun, and Yebin Liu.
\newblock Pymaf-x: Towards well-aligned full-body model regression from monocular images.
\newblock \emph{IEEE Transactions on Pattern Analysis and Machine Intelligence}, 2023.

\bibitem[Zhang et~al.(2025)Zhang, Deng, Ma, and Potamias]{zhang2025hawor}
Jinglei Zhang, Jiankang Deng, Chao Ma, and Rolandos~Alexandros Potamias.
\newblock Hawor: World-space hand motion reconstruction from egocentric videos.
\newblock In \emph{Proceedings of the Computer Vision and Pattern Recognition Conference}, pages 1805--1815, 2025.

\end{thebibliography}

\clearpage


\end{document}